\documentclass[conference]{IEEEtran}

\makeatother

\IEEEoverridecommandlockouts
\usepackage{amsmath,amssymb,amsfonts}
\usepackage{algorithmic}
\usepackage{graphicx}
\usepackage{textcomp}
\usepackage{xcolor}
\usepackage{url}
\usepackage{array}
\usepackage{adjustbox}
\usepackage{graphicx}
\usepackage{adjustbox}
\usepackage{algorithm}
\usepackage{amsmath}
\usepackage{hyperref}

\begin{document}

\title{Context Awareness Gate For Retrieval Augmented Generation}

% \IEEEpubid{\makebox[\textwidth]{979-8-3315-2225-4/24/\$31.00 ©2024 IEEE \hfill} \hspace{-\textwidth}\makebox[\textwidth]{ }}

\author{\IEEEauthorblockN{Mohammad Hassan Heydari}
\IEEEauthorblockA{\textit{Computer Engineering Faculty} \\
\textit{University of Isfahan}\\
Isfahan, Iran \\
mheydarii@mehr.ui.ac.ir}
\and
\IEEEauthorblockN{Arshia Hemmat}
\IEEEauthorblockA{\textit{Computer Engineering Faculty } \\
\textit{University of Isfahan}\\
Isfahan, Iran \\
arshiahemmat@mehr.ui.ac.ir}
\and
\IEEEauthorblockN{Erfan Naman}
\IEEEauthorblockA{\textit{Computer Engineering Facu.} \\
\textit{University of Isfahan}\\
Isfahan, Iran \\
erfannaman@mehr.ui.ac.ir}

\and
\IEEEauthorblockN{Afsaneh Fatemi}
\IEEEauthorblockA{\textit{Computer Engineering Faculty} \\
\textit{University of Isfahan}\\
Isfahan, Iran \\
a\_fatemi@eng.ui.ac.ir}

}

\maketitle

\begin{abstract}
Retrieval-Augmented Generation (RAG) has emerged as a widely adopted approach to mitigate the limitations of large language models (LLMs) in answering domain-specific questions. Previous research has predominantly focused on improving the accuracy and quality of retrieved data chunks to enhance the overall performance of the generation pipeline. However, despite ongoing advancements, the critical issue of retrieving irrelevant information—which can impair a model’s ability to utilize its internal knowledge effectively—has received minimal attention. In this work, we investigate the impact of retrieving irrelevant information in open-domain question answering, highlighting its significant detrimental effect on the quality of LLM outputs. To address this challenge, we propose the Context Awareness Gate (CAG) architecture, a novel mechanism that dynamically adjusts the LLM’s input prompt based on whether the user query necessitates external context retrieval. Additionally, we introduce the Vector Candidates method, a core mathematical component of CAG that is statistical, LLM-independent, and highly scalable. We further examine the distributions of relationships between contexts and questions, presenting a statistical analysis of these distributions. This analysis can be leveraged to enhance the context retrieval process in retrieval-augmented generation (RAG) systems \footnote{The code is publicly available at: \url{https://github.com/heydaari/CAG}}.
\end{abstract}


\section{Introduction}

Retrieval-augmented generation (RAG) has emerged as a leading approach for implementing question-answering systems that require intensive domain-specific knowledge \cite{lewis-rag}. This method allows for the utilization of customized datasets to generate answers, grounded in the information provided by those datasets. However, the effectiveness of the retrieval component within RAG pipelines is critical, as it directly influences the reliability and quality of the generated outputs \cite{chen,xiang}.

In efforts to enhance the quality of the retrieval component in RAG pipelines, research has demonstrated that transforming the user's input query into varying levels of abstraction before conducting the document search can significantly improve the relevance of the retrieved data. Several methods have been proposed, including query expansion into multi-query searches, chain of verification \cite{chain, cot}, pseudo-context search \cite{hyde} and query transformation \cite{rewriting, rewriting2, re3, survey}. These approaches contribute to more accurate and effective retrieval of information.

Despite ongoing efforts to develop more reliable retrieval methods for extracting relevant data chunks, many question-answering systems do not solely rely on local or domain-specific datasets for answering user queries. In addition to domain-specific user queries, many input queries do not necessitate retrieval from local datasets, which reduces the scalability and reliability of question-answering systems \cite{wang}. To tackle this limitation, retrieval methods based on query classification and routing mechanisms have proven effective in enhancing retrieval accuracy by directing the search toward a set of documents closely related to the user's query \cite{survey}. However, in our study, we demonstrate that even with semantic routing, the probability of retrieving irrelevant information remains non-negligible, particularly when dealing with a broad domain of potential queries.

Due to the inherently local search mechanism of Retrieval-Augmented Generation (RAG) systems \cite{lewis-rag, guu-rag}, even for queries that are largely irrelevant, the pipeline will still return a set number of passages. While existing research has made strides in addressing the challenge of imperfect data retrieval \cite{wang, prev-rare, prev-2}, the issue of broad-domain question answering in RAG systems has received relatively little attention.

Many queries submitted to RAG-enhanced question-answering (QA) systems do not require data retrieval, such as daily conversations, general knowledge questions, or questions that large language models (LLMs) themselves can answer using their internal knowledge \cite{survey, wang, best-prectice, adaptive}. Retrieving passages for all input queries, especially in these cases, can significantly diminish the retrieval precision \cite{wang, adaptive} and the context relevancy \cite{ragas}, often rendering them entirely irrelevant.

To address this issue, we propose a novel context-aware gate architecture for RAG-enhanced systems which is highly scalable of dynamically routing the LLM input prompt to increase the quality of pipeline outputs.

For better comprehension of our work, we highlight three main contributions in this study:

\begin{itemize}
\item \textbf{Context Awareness Gate (CAG)}: We introduce a novel gate architecture that significantly broadens the domain accessibility of RAG systems. CAG leverages both query transformation and dynamic prompting to enhance the reliability of RAG pipelines in both open-domain and closed-domain question answering tasks.

\item \textbf{Vector Candidates (VC)}: We propose a statistical semantic analysis algorithm that improves semantic search and routing by utilizing the concept of pseudo-queries and in-dataset embedding distributions.

\item \textbf{Context Retrieval Supervision Benchmark (CRSB) Dataset}: Alongside our technical and statistical investigations, we introduce the CRSB dataset, which consists of data from 17 diverse fields. We study the inner context-query distributions of this rich dataset and demonstrate the effectiveness and scalability of Vector Candidates on practical QA systems
\footnote{The CRSB dataset is available at: \url{https://huggingface.co/datasets/heydariAI/CRSB}} .
\end{itemize}

\begin{figure}[h]
  \centering
  \includegraphics[width=\linewidth]{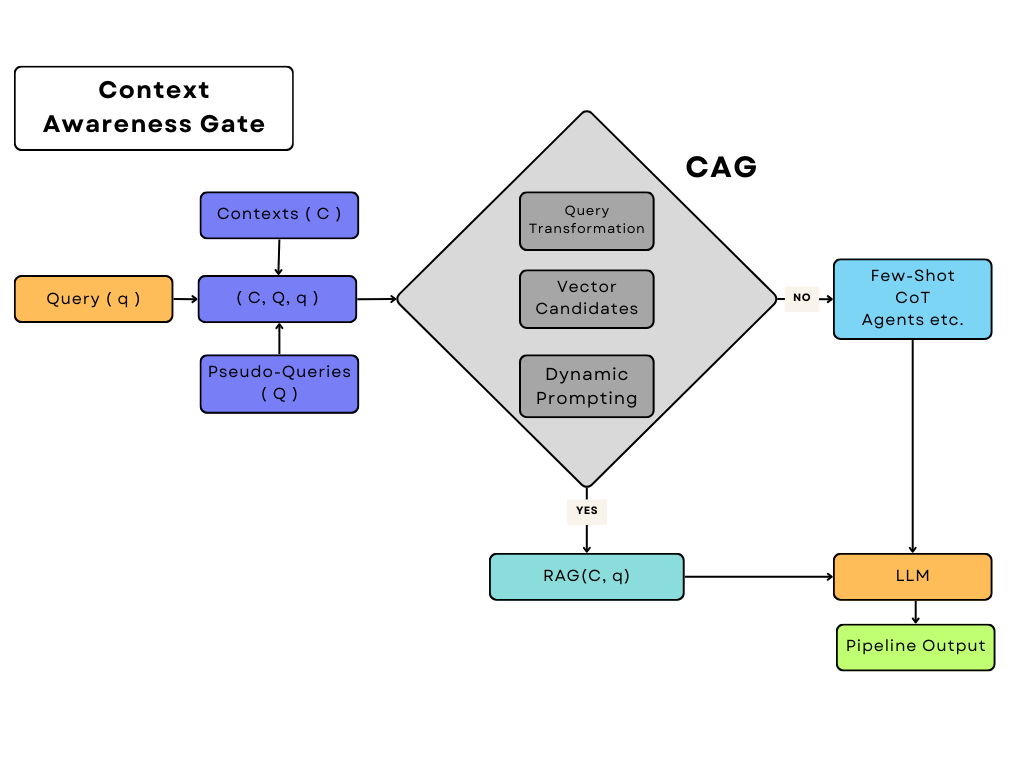} 
  \caption{Context Awareness Gate (CAG) architecture for open domain questions answering}
  \label{fig:figure1}
\end{figure}

\section{Related Work}

improving both retrieval quality and the outputs of large language models. Query2Doc \cite{query2doc} and HyDE \cite{hyde} generate pseudo-documents based on the input query and utilize these for semantic search instead of the query itself. RQ-RAG \cite{rq} decomposes complex queries into simpler sub-queries, enhancing retrieval performance. The Rewrite-Retrieve-Read framework \cite{rewriting} employs query rewriting to improve the match between queries and relevant documents. Additionally, some studies suggest that for queries answerable by the large language model (LLM) based on its internal knowledge, query classification using a smaller language model can benefit overall pipeline performance \cite{survey}.

In terms of improving model output quality, RobustRAG \cite{xiang} investigates the vulnerability of RAG-based systems to malicious passages injected into the knowledge database. Conflict-Disentangle Contrastive Decoding (CD2) \cite{cd2} proposes a framework to reconcile conflicts between an LLM's internal knowledge and external knowledge stored in a database. Yu et al. (2024) \cite{rankrag} argue that simply adding more context to the LLM input prompt does not necessarily improve performance. In a recent study, Wang et al. (2024) \cite{wang} show that when retrieval precision is below 20\%, RAG is not beneficial for QA systems. They highlight that when retrieval precision approaches zero, the RAG pipeline performs significantly worse than a pipeline without RAG. As a relevant method to our work, Adaptive-RAG \cite{adaptive} uses a smaller language model to adaptively switch the pipeline inputs whether it needs context retrieval for RAG or the question answering language model itself can answer the input query based on its internal knowledge.

\section{Approach}

To address the challenges associated with retrieving irrelevant information \cite{wang}, we propose the Context Awareness Gate (CAG) architecture, which utilizes Vector Candidates as its primary statistical method for query classification. CAG significantly improves the performance of open-domain question-answering systems by dynamically adjusting the input prompt for the LLM, transitioning from RAG-based context prompts to Few Shot, Chain-of-Thought (CoT) \cite{chain, cot}, and other methodologies. Consequently, the LLM responds to user queries based on its internal knowledge base.

\subsection{Context Awareness Gate (CAG)}
To address the issue of retrieving irrelevant data chunks for each input query, one solution is to ask a supervising large language model (LLM) to classify whether the query should prompt a retrieval-augmented generation (RAG) or a RAG-free response \cite{adaptive}. This involves determining whether the input query falls within the scope of the local database. However, a significant limitation of this approach is the high computational cost of using an LLM with billions of parameters for a relatively simple task like query classification. Even smaller language models come with their own challenges, such as hallucination and limited reasoning capabilities \cite{wang}.

To mitigate these issues,we propose an efficient yet highly effective statistical approach, known as Vector Candidates. The key idea behind Vector Candidates is to generate pseudo-queries for each document in the set, then calculate the distribution of embeddings and their similarities. By comparing the input query to this distribution, it is possible to estimate whether context retrieval is necessary with a certain level of probability. If the input query is far from this distribution, it is recommended not to retrieve any context and instead reformulate the LLM input prompt into a simpler few-shot question-answering task, rather than utilizing RAG. The overall architecture of CAG is presented in figure \ref{fig:figure1}

The limitation of this approach may appear to be the necessity of generating numerous pseudo-queries for a local dataset. However, when comparing it to LLM supervision \cite{adaptive}, the complexity of the Vector Candidates method, which operates on a set of contexts with \( C \) contexts and \( N \) pipeline input requests, reveals a significant advantage. Specifically, the complexity of the Vector Candidates method is \( O(1) \), as it relies solely on the number of contexts, regardless of the number of input requests ( This happens when we disable the query transformation as one of the CAG steps ) . In contrast, the complexity of LLM supervision \cite{adaptive} scales linearly with the number of input requests, represented as \( O(N) \). This indicates that while generating pseudo-queries may seem cumbersome, the overall efficiency of the Vector Candidates approach is superior in scenarios with multiple input requests.

Alongside all the steps involved in the Vector Candidates approach, the process begins with transforming the user's input query into a more appropriate format to enhance the quality of semantic search. This query transformation is critical as it ensures that the input is optimized for better alignment with the embeddings used in the retrieval process. After this transformation, the Vector Candidates method is applied to assess the relevance of context retrieval.

Following query transformation and Vector Candidates analysis, the Context Awareness Gate (CAG) system dynamically adapts the input query into a suitable prompt. This involves determining whether context retrieval is necessary or if the LLM can answer the query based on its internal knowledge, utilizing techniques like Chain of Thought (CoT) reasoning \cite{chain, cot}, agents, or other methods.

\subsection{Vector Candidates}
To address the issues of using a LLM for context supervision , we propose a statistical approach based on the distributions of emmbedings of contexts and pseudo-queries.

\begin{algorithm}
\caption{Vector Candidates Algorithm}
\label{alg:vector_candidates}
\begin{algorithmic}[1]
\REQUIRE Contexts $C$, Pseudo Queries $Q$, Policy $P$, Threshold $T$, Input Query $q$
\ENSURE A classification (True or False)

\STATE Compute dataset distributions based on cosine similarity:
\[
D \gets \frac{C \cdot Q}{\|C\| \|Q\|}
\]
\STATE Compute input query similarities with contexts:
\[
d \gets \frac{C \cdot q}{\|C\| \|q\|}
\]
\IF{$\max(d) > P(D) - T$}
    \RETURN True
\ELSE
    \RETURN False
\ENDIF

\end{algorithmic}
\end{algorithm}

Based on the proposed method in Algorithm \eqref{alg:vector_candidates}, we first calculate the cosine similarity distributions between the contexts and pseudo-queries. Then, we compute the similarity between the user's original query and each context in the dataset. If the maximum similarity found between the original query and the contexts falls within the distribution of context-pseudo-query similarities, this suggests that retrieval-augmented generation (RAG) might be beneficial. Otherwise, it is more efficient to exclude RAG from the pipeline. This approach is grounded in our statistical analysis and the results presented by Wang et al. \cite{wang}.

To measure the relevancy between the described distributions and the user query, we apply a policy \( P \), which is a hyperparameter derived from common statistical metrics such as minimum, mean, median, or quartiles. Additionally, we define a threshold \( T \), which serves as another hyperparameter, to create a risk range for decision-making. This threshold helps in determining the confidence level for whether context retrieval should be applied, balancing the trade-off between precision and recall in the retrieval process.

\subsection{Context Retrieval Supervision Bench (CRSB)}
We introduce the Context Retrieval Supervision Bench
(CRSB) dataset, which can be used to evaluate the per-
formance of context-aware systems and retrieval-augmented
generation (RAG) semantic routers. The CRSB contains 17
different topics, with each context associated with 3 pseudo-
queries. This design allows the CRSB to encompass a total of
5,100 question-answer pairs. With a correct permutation, CRSB can offer more than 83000 context-query pairs to evaluate the context awareness systems and semantic routing pipelines..

\section{Experiments}

To analyze the statistical relationships between relevant and irrelevant context-query pairs, we examine the distribution of collected contexts and generated pseudo-queries. We begin by gathering 1,700 contexts across 17 distinct topics. For each context, we prompt the Gemma 2 9B language model \cite{gemma} to generate three pseudo-queries.We applied \( all-mpnet-base-v2 \) as our embedding model and create a vector database of contexts and pseudo-queries embeddings \cite{sbert}.

We then calculate the similarity distributions for \textit{Positive} (relevant) context-query pairs, where the queries require context retrieval, as well as for \textit{Negative} (irrelevant) context-query pairs. With appropriate permutations, we analyze 83,000 \textit{Positive} and \textit{Negative} context-query pairs. Various statistical metrics are applied to these distributions, and the results are presented in Table \ref{tab:statistical_analysis}.

\begin{table}[ht]
\centering
\caption{Statistical Analysis on CRSB}
\begin{tabular}[t]{lcc}
\hline
\textbf{Policy} & \textbf{Positive} & \textbf{Negative} \\
\hline
\hline
\textbf{Minimum} & 0.110 & -0.193 \\
\textbf{5th Percentile} & 0.554 & -0.052 \\
\textbf{1st Quartile} & 0.662 & -0.000 \\
\textbf{Mean} & 0.705 & 0.047 \\
\textbf{Median} & 0.716 & 0.039 \\
\textbf{3rd Quartile} & 0.762 & 0.086 \\
\textbf{95th Percentile} & 0.836 & 0.219 \\
\textbf{Maximum} & 0.912 & 0.654 \\
\hline
\hline
\end{tabular}
\label{tab:statistical_analysis}
\end{table}

As demonstrated in Table \ref{tab:statistical_analysis}, over 95\% of positive context-question pairs exhibit a cosine similarity greater than 0.55, while more than 95\% of negative context-query pairs have a cosine similarity lower than 0.21. The median value for the positive set exceeds 0.71, whereas the median for the negative set is below 0.04. Although the maximum value of the negative set is higher than the minimum value of the positive set, the density of the positive distribution is greater than that of the negative distribution approximately 98.7\% of the time. These statistics provide a comprehensive understanding that, by utilizing these metrics as a policy, we can develop a statistical method that is highly effective in classifying user queries to establish dynamic prompts, as discussed in previous sections.

Due to the algebraic nature of our method, we have integrated advanced high-performance techniques for parallel computing and accelerated linear algebra through the JAX framework \cite{scaling}. Leveraging JAX's ability to handle automatic differentiation and just-in-time compilation (JIT) seamlessly, we are able to optimize the underlying computations for both CPU and GPU architectures. This not only allows for faster execution but also ensures scalability across large datasets and complex models. Our approach significantly improves the efficiency in computing the distributions of the dataset, offering a more streamlined and scalable solution for high-dimensional data analysis.

\section{Results}

To evaluate the capabilities and performance of the Context Awareness Gate (CAG), we applied this architecture to the SQuAD dataset \cite{rajpurkar2016squad} and our proposed benchmark, CRSB. We implemented an open-domain question-answering pipeline to assess the outcomes of CAG under a specified scenario:

\begin{itemize}
\item Setting CRSB as the local database while querying from SQuAD \cite{rajpurkar2016squad}. The pipeline should identify irrelevant queries to this local database and refrain from using RAG, instead generating a few-shot response using the LLM input prompt.

\end{itemize}

In this scenario, we evaluate the pipeline outputs using two metrics from RAGAS: context relevancy and answer relevancy \cite{ragas}.Due to the absence of retrieved context for irrelevant queries to the dataset, we ask our model to generate a pseudo-context that answers the query and then calculate the context relevancy based on this generated context. Our question-answering base model was Qwen-2.5-7B \cite{qwen2} and we applied Llama-3.2-3B as our answer relevancy evaluator \cite{llama}. To demonstrate the effectiveness of the proposed pipeline, we compare the results of the classic RAG \cite{lewis-rag}, HyDE \cite{hyde}, query transformation \cite{rewriting}, Adaptive-RAG \cite{adaptive} and the proposed CAG.

For the evaluation step, we applied both RAG and CAG. We set 95\% density distribution as the Policy \(P\) and we set the threshold \(T\) to 0 as the Vector Candidates hyperparameters.

\begin{table}[ht]
\centering
\caption{Evaluation of Context Awareness Gate (CAG) on SQuAD and CRSB}
\begin{tabular}[t]{lcc}
\textbf{ } & \textbf{Context Relevancy} & \textbf{Answer Relevancy} \\
\hline
\textbf{RAG \cite{lewis-rag}} & 0.016 & 0.206 \\
\textbf{Query Rewriting \cite{rewriting}} & 0.026 & 0.147 \\
\textbf{HyDE \cite{hyde}} & 0.043 & 0.228\\
\textbf{Adaptive-RAG \cite{adaptive}} & 0.334 & 0.613\\
\textbf{CAG (Ours)} & \textbf{0.338} & \textbf{0.709}\\
\hline
\end{tabular}
\label{tab:results}
\end{table}

Our experimental results shown in Table \ref{tab:results} clearly demonstrate that classic Retrieval-Augmented Generation (RAG) \cite{lewis-rag} is unable to generalize effectively for open-domain question answering. Both Query Rewriting \cite{rewriting} and HyDE \cite{hyde}, while providing improvements in context retrieval, share the same limitation as classic RAG \cite{lewis-rag} in failing to address the broader challenges posed by open-domain queries. As illustrated in our results, the context relevancy scores for these methods are close to zero (0.016 for RAG \cite{lewis-rag}, 0.026 for Query Rewriting \cite{rewriting}, and 0.043 for HyDE \cite{hyde}), indicating that they are unable to adapt the pipeline outputs in scenarios where the input query does not align with the local dataset. Consequently, their ability to generate accurate and relevant answers is also compromised, as reflected by their low answer relevancy scores (0.206 for RAG \cite{lewis-rag}, 0.147 for Query Rewriting \cite{rewriting}, and 0.228 for HyDE \cite{hyde}).

In contrast, Adaptive-RAG \cite{adaptive} and CAG (our proposed method) show a marked improvement in both context relevancy and answer relevancy. The results for Adaptive-RAG \cite{adaptive} ( context relevancy: 0.334, answer relevancy: 0.613) demonstrate its ability to adaptively decide when to trigger context retrieval based on query classification. Similarly, CAG achieves even higher performance (context relevancy: 0.338, answer relevancy: 0.709), suggesting that both methods successfully enhance the pipeline's ability to dynamically adjust the input to the language model, either by incorporating external context or by relying solely on the internal knowledge of the model, depending on the query's nature.

However, a close examination of the computational costs associated with each approach reveals key differences. While Adaptive-RAG \cite{adaptive} relies on an additional large language model (LLM) to supervise the decision of whether context retrieval is needed, this introduces significant overhead in terms of memory usage and inference time. As detailed in previous works, such LLM-based supervision requires additional model loading, which can considerably slow down the pipeline and increase resource consumption.

On the other hand, our CAG method offers a more efficient and scalable solution by eliminating the need for LLM supervision. Instead, CAG uses a statistical approach based on the distributions of the local database, allowing it to make context retrieval decisions in a highly efficient manner. This method is LLM-free and leverages a lightweight, statistical computation that can be hundreds or even thousands of times faster than LLM-based adaptations, offering significant advantages in both speed and scalability without compromising accuracy.

Thus, the results not only demonstrate the effectiveness of CAG in addressing the challenges of open-domain question answering, but also highlight its computational efficiency, making it a highly scalable solution for practical deployments of RAG-based systems.

\section{Future Direction}
This work opens up several promising avenues for further research and enhancement of Context Awareness Gate (CAG) in open-domain question answering systems:

\begin{itemize}
\item \textbf{Incorporating Best Practices in Information Retrieval}: Future work could focus on integrating the methodologies outlined in \cite{best-prectice} to refine CAG’s information retrieval pipeline. Specifically, these practices could optimize how CAG filters and ranks relevant information, leading to even more precise data selection. By enhancing the granularity of relevance scoring during retrieval, CAG could further improve its ability to identify and utilize the most contextually pertinent chunks of information, boosting both retrieval accuracy and downstream performance in generating high-quality responses.

\item \textbf{Replacing Pseudo-Context Search with Pseudo-Query Search}: While the pseudo-context search strategy proposed in HyDE \cite{hyde} has been effective, this study introduces the concept of pseudo-query search as a potentially more robust alternative. Future research could explore the efficacy of this approach across various datasets and domains. A systematic evaluation of the pseudo-query search could reveal whether it generalizes better across different question-answering tasks, especially in complex or multi-turn dialogues, where context awareness is crucial for maintaining conversation coherence.

\end{itemize}

\bibliographystyle{IEEEtran}
\bibliography{IEEEabrv, refrences}

\end{document}